\documentclass{article}

\usepackage[preprint]{neurips_2026}

\usepackage{amsmath,amssymb}
\usepackage[utf8]{inputenc} 
\usepackage[T1]{fontenc}    
\usepackage{hyperref} 
\usepackage{cleveref}  
\usepackage{url}            
\usepackage{booktabs}       
\usepackage{amsfonts}       
\usepackage{nicefrac}       
\usepackage{microtype}      
\usepackage{xcolor}       
\usepackage{xspace}
\usepackage{enumitem}
\usepackage{adjustbox}
\usepackage{mathtools}
\usepackage{wrapfig}
\usepackage{tikz}
\usetikzlibrary{positioning}

\newcommand{\datasetsize}{34K\xspace}

\makeatletter
\newcommand*{\bigcdot}{}
\DeclareRobustCommand*{\bigcdot}{%
  \mathbin{\mathpalette\bigcdot@{}}%
}
\newcommand*{\bigcdot@scalefactor}{.5}
\newcommand*{\bigcdot@widthfactor}{1.15}
\newcommand*{\bigcdot@}[2]{%
  \sbox0{$#1\vcenter{}$}
  \sbox2{$#1\cdot\m@th$}%
  \hbox to \bigcdot@widthfactor\wd2{%
    \hfil
    \raise\ht0\hbox{%
      \scalebox{\bigcdot@scalefactor}{%
        \lower\ht0\hbox{$#1\bullet\m@th$}%
      }%
    }%
    \hfil
  }%
}
\makeatother

\title{\name: Language-Driven Cinematographic Framing for Human-Centric Video Generation}

\author{%
  M. Burak Kizil \\
  Ko\c{c} University\\
  \And
  Enes Sanli \\
  Ko\c{c} University \\
  \And
  Niloy J. Mitra \\
  University College London, Adobe  \\
  \And
  Xuelin Chen \\
  Adobe  \\
  \And
  Erkut Erdem \\
  Hacettepe University \\
  \And
  Aykut Erdem \\
  Ko\c{c} University \\
  \And
  Duygu Ceylan \\
  Adobe  \\
}
\newcommand{\name}{Auteur\xspace}

\begin{document}

\maketitle

\begin{abstract}
Generative video models have achieved remarkable visual fidelity and temporal coherence, yet intentional camera control remains elusive. Existing frameworks treat camera motion as a byproduct of pixel synthesis, producing trajectories that are stochastic, spatially inconsistent, and indifferent to the human subject driving the scene.
In this work, we present \name, a method for language-driven, human-centric camera \textit{framing} in generative video. Our core insight is that professional filmmakers conceive shots not as world-space trajectories but as framings defined relative to the actor, encoding shot size, angle, and composition as functions of human pose and motion. We formalize this intuition as a human-centric camera parameterization and introduce a Domain-Specific Language~(DSL) that is convertible to standard 6-DoF camera parameters. A fine-tuned multimodal large language model then acts as a virtual \textit{director}, mapping natural language descriptions and coarse human motion to sparse DSL keyframes that are deterministically interpolated into continuous camera trajectories, which are then provided as input to video generators. We train and evaluate \name\ on a new dataset of \datasetsize aligned text, human motion, and DSL-annotated camera trajectories drawn from procedural synthesis and real-world movie footage from the CondensedMovies \citep{condensedMovies} dataset.
\name enables cinematographic framing of human-centered scenes, a capability largely absent in prior generative  models. To assess this behavior, we propose new framing-focused metrics, and our experiments show that \name consistently outperforms existing methods.
\end{abstract}

\begin{flushright}
    \begin{minipage}{0.7\textwidth}
        \itshape 
        “There are no good and bad movies, only good and bad directors.” \\
        \raggedleft \normalfont --- \textsc{François Truffaut}
    \end{minipage}
\end{flushright}

\section{Introduction}
The essence of cinematography lies not in simply recording a scene but in intentionally orchestrating perspective to evoke narrative meaning. While recent generative video models~\citep{kong2024hunyuanvideo,openai2024sora,wan2025,veo3_2026} have made remarkable strides in visual fidelity and temporal coherence, they lack the structured, deliberate camera behavior required to tell a coherent visual story. Current methods for camera control in video generation enable direct conditioning on explicit 6-DoF geometric trajectories~\citep{cameractrl2,Cheong_2025_BMVC} but specify them in world-space coordinates, decoupled from the actors on-screen. Often, transferring a trajectory extracted from reference footage to a new scene produces shots that are geometrically plausible but narratively meaningless; the camera moves, but not in response to the subject's performance. Object-centric approaches~\citep{LAMP} and sampling-based alternatives~\citep{PulpMotion} reduce subject-frame violations but still do not capture the deliberate synchronization between actor dynamics and camera response that defines cinematography.

Directors of photography do not conceive of shots as mere raw spatial trajectories~\citep{mascelli1965five,Katz1991,Bordwell2020}. Instead, they reason in terms of \emph{framing}: shot size (close-up, medium, wide), camera angle (eye-level, low, overhead), and screen composition (lead room, headroom), all defined relative to the actor’s body and movement over time. Cinematography is therefore not an independent geometric construct, but a \emph{function} of the subject's position, movement, and narrative role within the scene. We argue that a generative system capable of authentic cinematography must
transition from a passive observer to an active \textit{auteur}: an agent that deliberatively \emph{authors} camera behavior with awareness of the humans and their space-time choreography. 


\begin{figure}[!t]
    \centering
\includegraphics[width=\linewidth]{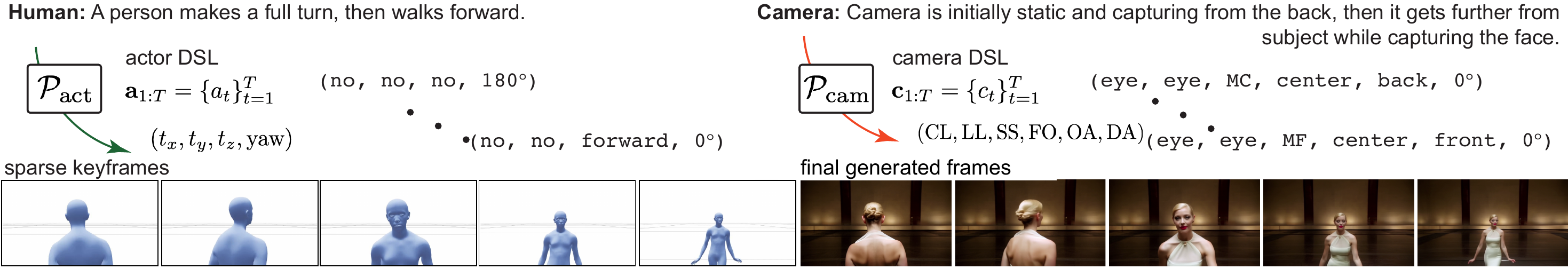}
    \caption{\textbf{\name\ defines every camera decision relative to the human subject:} how much of the body is visible (\textit{scale}), from which angle the actor is viewed (\textit{orientation}), and how they are placed in the frame (\textit{composition}). Given a natural-language description (top), a fine-tuned multimodal LLM generates a structured DSL program (both actor and camera DSLs) that encodes these actor-relative parameters at sparse keyframes, which are then deterministically interpolated into a continuous 6-DoF camera trajectory conditioned on the human mesh (blue). The trajectory then guides video synthesis.}
    \label{fig:placeholder}
\end{figure}

Building on this insight, we introduce \name, a method for composing human-aware camera motion from natural language.
Our approach is grounded in three key components: a human-centric camera representation, a two-stage generative formulation, and a dataset that anchors both in real or synthetic footage.
At the core of \name is a \textit{human-centric camera parameterization} that represents camera state along semantic axes, namely orientation, shot scale, camera level, framing, and look-at level, all defined relative to the actor's body coordinate frame derived from SOMA~\citep{saito2026soma}. We implement this continuous parameterization as a \textit{Domain-Specific Language}~(DSL): a structured, discrete form that is human-readable and interpretable, generated by a large language model, and can be deterministically converted to standard 6-DoF camera parameters. The DSL can be viewed as a \emph{quantization} of the underlying parameterization, mapping a continuous space into a vocabulary of cinematographically meaningful values while preserving full geometric expressiveness.



Building upon this representation, \name\ operates in two stages. First, a fine-tuned multimodal LLM, Qwen-2.5-VL~\citep{qwen2.5-VL} maps a natural language description to a coarse 3D human trajectory, capturing the subject's spatial evolution. Second, conditioned on this trajectory, the LLM acts as a virtual director, generating DSL programs that specify camera state at sparse, scene-defining keyframes, such as shot initiations and transitions. These keyframes are deterministically interpolated into a continuous camera trajectory that is both geometrically consistent and cinematographically meaningful, and is used to condition downstream video generation models.

To train \name, we construct a dataset combining sythetic procedural and real-world footage. For procedural data, we synthesize scenes by pairing randomized SOMA~\citep{saito2026soma} motion sequences with DSL-driven camera programs, rendering them via a 3D engine to obtain (human motion, camera trajectory, DSL program) triplets. For real-world footage, we process videos from CondensedMovies~\citep{condensedMovies} to recover 4D human motion and camera trajectories~\citep{tram}; captions are generated with AuroraCap~\citep{chai2024auroracap}, and DSL annotations are obtained via our motion-tagging strategy. This yields \datasetsize aligned (caption, 3D human motion, camera trajectory, DSL program) tuples that ground our learning of human-aware cinematic priors in real filmmaking practice.
%
%
We supply the generated camera trajectories by \name to multiple downstream video generation models, demonstrating improved spatial coherence, subject alignment, and perceived cinematic quality over prior methods.
%
In summary, our contributions are:
\begin{itemize}[leftmargin=*]
\item \textbf{Human-centric camera parameterization and DSL.} An actor-relative representation of camera state grounded in professional cinematographic conventions, with \emph{framing}, rather than raw trajectory, as the primary compositional primitive. The DSL is its discrete LLM-generatable operationalization, deterministically convertible to standard 6-DoF parameters.
 
 \item \textbf{Language-to-human-to-camera pipeline.}
  A two-stage method mapping natural language to coarse 3D human trajectories, then to sparse DSL keyframe programs that are deterministically interpolated into continuous, actor-aware 6-DoF camera paths.
\item \textbf{\name\ dataset.}
  A dataset of \datasetsize (caption, SOMA parameters over time, camera trajectory, DSL program) samples from  real data, enabling learning of human-aware cinematic priors. 
\item \textbf{Comprehensive cross-dataset evaluation}
We introduce \name Score, a new framing-centric benchmark suite measuring subject visibility, compositional adherence, temporal framing stability, and actor-camera coordination. Across both our internal Auteur benchmark and the external PulpMotion benchmark, our method outperforms prior controllable camera baselines \citep{LAMP, PulpMotion}, achieving stronger controllability, more stable cinematic framing
\end{itemize}
\section{Related Work}

\paragraph{Explicit camera conditioning in generative video.}
Large-scale video diffusion models (e.g., Sora, Wan, Veo) produce compelling photorealistic visual priors of scene content but treat viewpoint as an uncoupled, passive byproduct of the sampling process.
Recent work inject user-specified camera signals through a range of geometric interfaces: raw extrinsics~\citep{motionctrl}, Plücker embeddings~\citep{cameractrl2}, dense point trajectories~\citep{i2vcontrol,realcam,gokmen2025ropecraft,vidcraft3,flovd:25}, and persistent 3D geometry~\citep{gen3c,epic,agarwal2025cosmos,cinemaster}. A complementary line uses point-cloud renderings or multi-camera setups to enforce geometric adherence~\citep{camtrol,trajectorycrafter,recammaster,vista4d,versecrafter}. While these methods substantially improve viewpoint controllability, they all treat the camera trajectory as an externally authored, independent variable. As a result, they function as conditional renderers $p(\text{video}\mid\text{camera},\text{text})$ rather than addressing the prior question of \emph{how} a camera should move to meaningfully frame the actors and their actions. \name instead generates the trajectory itself, conditioned on the actors' physical state, which can then be fed into camera-conditioned
generators such as those above.
\vspace{-2mm}
\paragraph{Foundations of cinematic grammar.}
Classical film theory formalizes visual storytelling as a structured grammar. Seminal works codify a precise vocabulary of shot scales and framing constraints~\citep{mascelli1965five, mercado2010filmmaker} and articulate rules for actor staging, blocking, and shot-by-shot directorial planning~\citep{Katz1991, Bordwell2020}. Our DSL (\Cref{sec:representation}) operationalizes this vocabulary as a parameterized, machine-generatable representation, bridging abstract directorial intent and programmable continuous camera control.
\vspace{-2mm}
\paragraph{Framing-aware planning and virtual cinematography.}
Prior work inspired by classical virtual cinematography has long formalized screen-space composition, visibility, and shot design through methods (e.g. Through-the-Lens control~\citep{throughthelens} and Toric-space parameterizations~\citep{toricspace}). These are later extended to drone cinematography as a constrained optimization over smoothness, occlusion, and target framing~\citep{drone_camera:18}. In the generative setting, E.T.~\citep{courant2024et} and GenDoP~\citep{zhang2025gendop} learn camera paths from text and character trajectories, \citet{cheng_automating_visual:25} refine paths via Toric interaction features, and the concurrent VERTIGO~\citep{vertigo:26} post-trains a camera generator against learned visual preferences. Most closely related to ours, PulpMotion~\citep{PulpMotion} introduces an auxiliary framing latent derived from screen-space projected joints to encourage coherence. These methods tie camera behavior to actors or cinematic heuristics, but each derives the camera from text, discrete rules, auxiliary sampling, or 2D screen-space proxies, but never from a parameterized 3D body-space representation. A separate line of work~\citep{CCD, Director3D} jointly generates camera and video without conditioning on actor performance, while LLM-based scene planners~\citep{modularcam, Lin2023VideoDirectorGPT, song2025llamalearnsdirectdirectorllm} decompose text into multi-scene prompts that downstream generators render, again without modeling the camera as a response to actor motion. While the recent LAMP~\citep{LAMP} work models camera tracking behavior from natural language, it does not handle the explicit framing of the camera, which is the gap \name addresses.
\vspace{-2mm}
\paragraph{Actor-aware camera movement.}
Existing joint human–camera systems~\citep{humanvid:24, cao2025uni3c} model both entities but largely expose the camera as a parallel, decoupled control (input) stream or rely on external descriptors for coupling. We instead formulate cinematography as an actor-conditioned framing task. By operating directly over a unified parametric body manifold via SOMA~\citep{saito2026soma}), \name learns camera behavior as a conditional distribution $p(\text{camera} \mid \text{action})$, where the action is the SOMA-parameterized space-time trajectory of the actor. This shifts the paradigm from treating the camera as a geometric input or text-derived heuristic to a learned policy that responds to the subject and is consumable by downstream video generators.












\section{Method}
\begin{figure}[t!]
    \centering
\includegraphics[width=\linewidth]{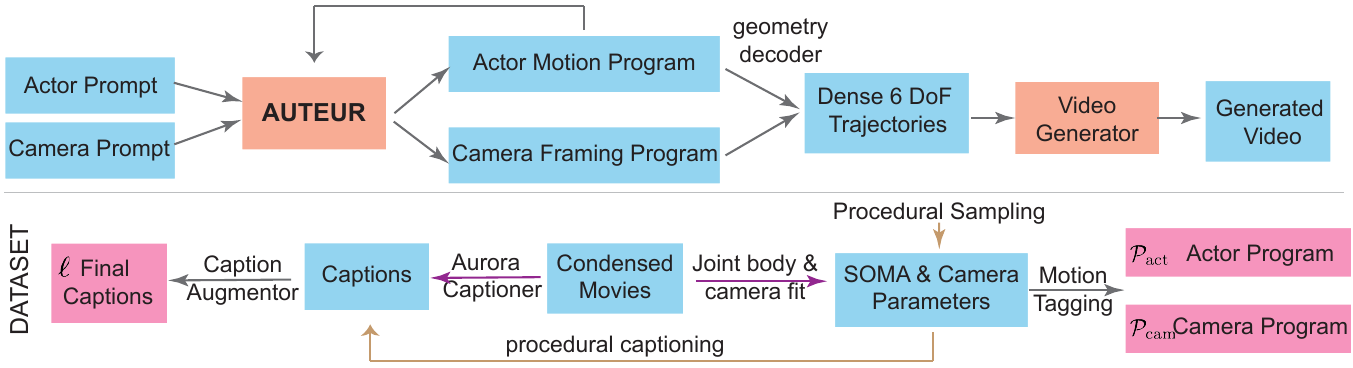}
    \caption{\textbf{Overview of \name.} Given a text prompt, \name predicts a coarse actor motion program and a camera framing program. The camera program specifies sparse actor-relative framing keyframes in our DSL. These are embedded interpolated over time, and decoded into 6-DoF trajectories that can be used for conditioning video generators. \name is trained with caption, actor, and camera program tuples $(\ell,\, \mathcal{P}_{\mathrm{act}},\, \mathcal{P}_{\mathrm{cam}})$ obtained procedurally and from real footage~\cite{condensedMovies}.}
    \label{fig:overview}
\end{figure}

\paragraph{Overview.}
\name is a two-stage framework that maps free-form natural language description $\ell$ to (i) a coarse 3D actor trajectory $\mathbf{a}_{1:T} = \{a_t\}_{t=1}^{T}$ and (ii) an actor-aware 6-DoF camera trajectory $\mathbf{c}_{1:T} = \{c_t\}_{t=1}^{T}$ suitable for conditioning downstream video generators. At the core of our approach is a \emph{human-centric camera parameterization} (\Cref{sec:representation}) that expresses the camera state at each frame in the actor's corresponding local body frame, together with a \emph{cinematographic Domain-Specific Language} (DSL) that discretizes this parameterization into a compact, LLM-generatable program. Sparse DSL keyframes are deterministically interpolated into a dense camera trajectory (\Cref{app:interp}). A fine-tuned multimodal LLM acts as a \emph{virtual director} (\Cref{sec:planner}): it first generates the actor trajectory and then, conditioned on it, emits the sparse DSL keyframes. The resulting trajectory drives a range of camera-conditioned video generators (\Cref{sec:videogen}). \Cref{fig:overview} summarizes the pipeline.

\subsection{Human-Centric Camera Parameterization}
\label{sec:representation}
\paragraph{Why not world-space 6-DoF?} 
World-space $\mathrm{SE}(3)$  trajectories are the de-facto interface for camera-conditioned video models, yet they are an unnatural medium for cinematographic intent. A trajectory recorded around one actor and replayed around another produces shots that are geometrically valid but narratively misfit; the camera moves through space, but not \emph{with respect to} the subject's pose, facing, or action. Professional filmmakers instead reason and operate in terms of \emph{framing}, which include shot scale, camera angle, and screen composition, all defined relative to the actor's body~\citep{mascelli1965five,Katz1991,Bordwell2020}. 

\paragraph{Actor state and body frame.}
We model the actor at frame $t \in \{1,\ldots,T\}$ as, 
\begin{equation}
a_t := \big(\mathbf{p}_t,\; \psi_t,\; h_t\big) \in \mathbb{R}^3 \times \mathbb{S}^1 \times \mathbb{R}_{>0},
\label{eq:actor-state}
\end{equation}
where $\mathbf{p}_t$ is the pelvis position, $\psi_t$ is the body yaw (facing angle in the ground plane), and $h_t$ is the body height in meters; all derived from the SOMA parametric body model~\citep{saito2026soma} in A-pose. We define the actor-anchored coordinate frame as, 
\begin{equation}
\mathcal{F}_t := \big(\mathbf{p}_t,\; \hat{\mathbf{f}}_t,\; \hat{\mathbf{r}}_t,\; \hat{\mathbf{z}}\big),
\qquad
\hat{\mathbf{f}}_t = (\cos\psi_t, \sin\psi_t, 0),
\qquad
\hat{\mathbf{r}}_t = \hat{\mathbf{z}} \times \hat{\mathbf{f}}_t,
\end{equation}
with $\hat{\mathbf{z}}$ the world up-axis. All camera parameters are expressed in this actor-centric local $\mathcal{F}_t$. 


\paragraph{Framing state.} 
A camera state is represented by six actor-relative variables. We let $\mathcal{A} := \{\text{CL}, \text{LL}, \text{SS}, \text{FO}, \text{OA}, \text{DA}\}$ index the camera axes; each axis $a \in \mathcal{A}$ takes values in a continuous set $\mathcal{K}_a$ summarized in Table~\ref{tab:axes}. 
They are: \\
(i)~\textbf{CL}\,(Camera Level) $\in \mathcal{K}_{\text{CL}} = \mathbb{R}_{\geq 0}$.\;Camera height above the ground plane.\\
(ii)~\textbf{LL}\,(Look-at Level) $\in \mathcal{K}_{\text{LL}} = [0, h_t]$.\;Height on the actor at which the optical axis is aimed. \\
(iii)~\textbf{SS}\,(Shot Scale) $\in \mathcal{K}_{\text{SS}} = (0, 1]$.\;Fraction of frame height occupied by the actor's body.\\
(iv) \textbf{FO}\,(Framing Offset) $\in \mathcal{K}_{\text{FO}} = [-\tfrac{1}{2}, +\tfrac{1}{2}]$.\;Lateral placement of the actor in normalized image coordinates, capturing
        lead-room and rule-of-thirds composition conventions.\\
(v)~\textbf{OA}\,(Orientation Angle) $\in \mathcal{K}_{\text{OA}} = \mathbb{S}^1$.\;Camera azimuth around the actor, measured from $\hat{\mathbf{f}}_t$.\\
(vi) \textbf{DA}\,(Dutch Angle) $\in \mathcal{K}_{\text{DA}} = \mathbb{S}^1$.\;Camera roll about the optical axis, encoding stylistic tilt.

The joint continuous state space is the product manifold
\begin{equation}
\mathcal{K} \;=\; \mathcal{K}_{\text{CL}} \times \mathcal{K}_{\text{LL}} \times \mathcal{K}_{\text{SS}} \times \mathcal{K}_{\text{FO}} \times \mathcal{K}_{\text{OA}} \times \mathcal{K}_{\text{DA}},
\end{equation}
a smooth 6-dimension manifold. Unlike world-space trajectories, this space describes \emph{how the actor should be framed}; the 6-DoF camera pose, $\mathbf{k}_t \in \mathcal{K}$, arises as a consequence of this framing decision. 


\paragraph{Cinematographic DSL representation.} 
The continuous spaces in which the camera framing state and the actor trajectory live are geometrically expressive but unsuitable as direct language-model outputs. Token-level regression of real numbers is brittle, and continuous predictions discard the inductive bias of an established cinematographic vocabulary. We therefore introduce a unified \emph{Domain-Specific Language} (DSL) that quantizes both spaces along semantic axes into compact, human-readable, LLM-generatable 
programs.

Formally, given a finite axis set $\mathcal{A}$ with per-axis continuous domains $\{\mathcal{K}_a\}_{a \in \mathcal{A}}$ and finite vocabularies $\mathcal{V}_a \subset \mathcal{K}_a$ of cinematographic tokens (e.g., $\mathcal{V}_{\text{SS}} = \{\textsc{ECU}, \textsc{CU}, \ldots, \textsc{EWS}\}$ for shot scale; full vocabularies in Appendix~\ref{app:dsl}), a DSL  \emph{program} is a sparse ordered sequence of \emph{shot records},
\begin{equation}
\mathcal{P} \;=\; \big((\tau_k, \delta_k)\big)_{k=1}^{K},
\qquad 1 \leq \tau_1 < \cdots < \tau_K \leq T,
\label{eq:dsl-program}
\end{equation}
where $\tau_k$ is a keyframe index and $\delta_k$ is a \emph{partial assignment}
\begin{equation}
\delta_k \in \Delta\mathcal{V} \;:=\; \bigsqcup_{S \subseteq \mathcal{A}} \prod_{a \in S} \mathcal{V}_a,
\label{eq:partial-assignment}
\end{equation}
specifying new values only for the axes $\mathrm{dom}(\delta_k) \subseteq \mathcal{A}$ that change at $\tau_k$. The carry-forward rule
\begin{equation}
\tilde{\mathbf{k}}_{\tau_k}[a] \;=\; 
\begin{cases}
\delta_k[a] & a \in \mathrm{dom}(\delta_k), \\[2pt]
\tilde{\mathbf{k}}_{\tau_{k-1}}[a] & \text{otherwise},
\end{cases}
\label{eq:carry-forward}
\end{equation}
extends each partial assignment to a fully specified discrete state, encoding the cinematographic convention that a director specifies only what \emph{changes} at each transition. A per-axis lookup 
$\varphi_a : \mathcal{V}_a \hookrightarrow \mathcal{K}_a$ embeds tokens into continuous values, and axis-aware spline interpolation (Appendix~\ref{app:interp}) densifies the keyframe sequence into a 
per-frame trajectory in the underlying continuous space.

\paragraph{Two instantiations.}
We instantiate this DSL machinery twice in Auteur:
\begin{itemize}[leftmargin=*]
\item \textbf{Camera DSL} ($\mathcal{P}_{\text{cam}}$) over axes $\mathcal{A}_{\text{cam}}$ defined above. After carry-forward decoding, embedding via $\varphi$, and interpolation, we obtain a dense framing state $\mathbf{k}_t \in \mathcal{K}_{\text{cam}}$ at every frame, which the geometric decoder $\Phi$ (Appendix~\ref{app:interp}) maps to a 6-DoF pose $(\mathbf{R}_t, \mathbf{t}_t) \in \mathrm{SE}(3)$ conditioned on the actor state $a_t$.
\item \textbf{Actor DSL} ($\mathcal{P}_{\text{act}}$) over axes $\mathcal{A}_{\text{act}} = \{\text{P}_x, \text{P}_y, \text{P}_z, \Psi\}$ encoding the pelvis position $\mathbf{p}_t$ and yaw $\psi_t$ from Eq.~\eqref{eq:actor-state} at sparse keyframes. Body height $h_t$ is treated as a clip-level constant outside the DSL. The same carry-forward and interpolation pipeline produces the dense per-frame actor state $a_{1:T}$ that conditions $\Phi$.
\end{itemize}

\paragraph{Three representational layers.}
\name factors generation through three layers. The director LLM emits discrete DSL tokens in $\mathcal{V}$. The per-axis lookup $\varphi$ embeds these tokens into the continuous space $\mathcal{K}$, where axis-aware spline interpolation produces dense per-frame trajectories. For the camera branch, the geometric decoder $\Phi(\,\cdot\,;\,a_t)$ then maps each framing state to a world-space pose in $\mathrm{SE}(3)$, conditioned on the actor state. Both $\varphi$ and $\Phi$ are deterministic and closed-form, so the only learned component in the pipeline is the LLM-based director.


\subsection{LLM-Based Director}
\label{sec:planner}
\paragraph{Two-stage autoregressive director.}
We fine-tune Qwen-2.5-VL~\citep{qwen2.5-VL} as a policy $\pi_\phi$ that, given a free-form text description $\ell$, sequentially generates the two DSL programs introduced before: the actor program $\mathcal{P}_{\text{act}}$ and the camera program $\mathcal{P}_{\text{cam}}$. Tokens within each program are emitted under aa grammar-constrained decoding scheme that guarantees syntactically valid DSL output (Appendix~\ref{app:dsl}). The actor-first ordering reflects the cinematographic principle that subject motion defines the dramatic context to which the camera responds; we discuss this choice below.

\paragraph{Factorization.}
Recall that each program is a sparse sequence of shot records $(\tau_k, \delta_k)$, with $\tau_k$ a
keyframe index and $\delta_k$ a partial assignment over the program's axis set. The joint distribution over both programs given the prompt factors as
\begin{equation}
p_\phi(\mathcal{P}_{\text{act}}, \mathcal{P}_{\text{cam}} \mid \ell)
\;=\;
\underbrace{\prod_{k=1}^{K_{\text{a}}} p_\phi\!\Big(\big(\tau_k^{\text{a}}, \delta_k^{\text{a}}\big) \,\Big|\, \mathcal{P}_{\text{act}}^{<k},\, \ell\Big)}_{\text{actor planning}}
\;\cdot\;
\underbrace{\prod_{k=1}^{K_{\text{c}}} p_\phi\!\Big(\big(\tau_k^{\text{c}}, \delta_k^{\text{c}}\big) \,\Big|\, \mathcal{P}_{\text{cam}}^{<k},\, \mathcal{P}_{\text{act}},\, \ell\Big)}_{\text{camera planning}},
\label{eq:factorization}
\end{equation}
where $\mathcal{P}_{\bigcdot}^{<k}$ denotes the prefix of records strictly before index $k$ in program $\mathcal{P}_{\bigcdot}$. Each record-level conditional further decomposes into a token-level autoregression over the DSL vocabulary $\mathcal{V}_{\bigcdot}$. This actor-first factorization encodes the causal structure of cinematography; the actor's coarse trajectory defines the scene's dramatic geometry, and the camera is planned as a deliberate response to it. We validate this causally grounded design in the ~\ref{sec:actor}, confirming it maintains predictive accuracy while remaining strictly necessary for decoding the actor-relative representation.

\paragraph{Training objective.}
Given supervised tuples $(\ell, \mathcal{P}_{\text{act}}, \mathcal{P}_{\text{cam}}) \sim \mathcal{D}$ from the dataset described in Section~\ref{sec:datasets}, we minimize the standard token-level negative log-likelihood induced by 
Eq.~\eqref{eq:factorization},
\begin{equation}
\mathcal{L}(\phi) \;=\; -\,\mathbb{E}_{\mathcal{D}}\Big[\, \log p_\phi(\mathcal{P}_{\text{act}} \mid \ell) \;+\; \log p_\phi(\mathcal{P}_{\text{cam}} \mid \mathcal{P}_{\text{act}}, \ell) \,\Big].
\end{equation}
Full training schedule and settings are reported in Appendix~\ref{app:training}.

\paragraph{Editability and human-in-the-loop control.}
Because $\mathcal{P}_{\mathrm{cam}}$ is a symbolic program, users can directly inspect and modify the generated shot plan before downstream video synthesis. For example, overriding a single keyframe from \textsc{MS} to \textsc{CU} changes the intended shot scale while leaving  every other axis and every other keyframe untouched. This provides interpretable control that is difficult to obtain from models that regress trajectories directly into $\mathrm{SE}(3)$.

\subsection{Interface to Video Generators}
\label{sec:videogen}
The output of \name is a dense actor trajectory $
\{ a_t \}_{t=1}^{T}
$ and a dense 6-DoF camera trajectory $\{(\mathbf{R}_t, \mathbf{t}_t)\}_{t=1}^{T}$ that is agnostic to the downstream generator. To demonstrate this generality, we integrate \name with two architectures spanning distinct conditioning modalities (\Cref{fig:overview}).

(i)~\textbf{Image-to-video via VerseCrafter.}
VerseCrafter~\citep{versecrafter} is a video diffusion transformer that accepts explicit camera signals. We encode the predicted extrinsics as Pl\"ucker embeddings~\citep{cameractrl2} and inject them into the spatial-attention layers. In addition, we extract a 3D Gaussian scene representation from the input image and warp it according to both the actor trajectory $\mathbf{a}_{1:T}$ and the camera trajectory, providing an explicit geometric scaffold for multi-view consistency.
(ii)~\textbf{Text-to-video via Kimodo + VACE}
Given the coarse actor trajectory produced by \name, Kimodo~\citep{kimodo2026} augments it with full articulated 3D human motion. We then render the resulting body mesh and we use VACE~\citep{vace} by conditioning on it through a rendered control video: at each frame, we project the actor under the predicted camera and composite it onto a blank canvas. This visual control signal implicitly encodes framing and perspective without requiring any architectural modification to the generator.

Across these architectures, human-centric trajectories produced by \name impose actor-aware framing constraints throughout the generated clip, directly addressing the subject--frame drift that afflicts purely geometric baselines (\Cref{sec:experiments}).



\subsection{Dataset}
\label{sec:datasets}

Training Auteur requires aligned tuples $(\ell,\, \mathcal{P}_{\mathrm{act}},\, \mathcal{P}_{\mathrm{cam}})$ of a natural-language caption, an actor DSL program, and a camera DSL program. We construct two complementary splits and combine them for training; full pipeline details and statistics are deferred to Appendix~\ref{app:dataset}.

\paragraph{Procedural data.} We synthesize $N_{\mathrm{proc}}$ tuples by pairing randomized SOMA~\citep{saito2026soma} motion sequences with camera programs sampled from our DSL grammar, decoded into ground-truth trajectories via $\Phi$ (Appendix~\ref{app:interp}). Captions are produced from structured templates that verbalize the sampled motion and framing parameters. This split affords full distributional control and near-uniform coverage of all DSL axes, including rare multi-axis transitions.

\paragraph{Real-world data.} We mine $N_{\mathrm{real}}$ tuples from CondensedMovies~\citep{condensedMovies} by recovering metric-scale 3D human and camera trajectories with a TRAM-style estimator~\citep{tram}, projecting them into our human-centric parameter space (Section~\ref{sec:representation}), tagging scene-defining moments with the nearest DSL tokens to obtain $\mathcal{P}_{\mathrm{cam}}$, and captioning each clip with AuroraCap~\citep{chai2024auroracap} (augmented by prompt rewriting~\citep{geminiteam2024gemini15}).

\paragraph{Why both.} The two splits are complementary: procedural data provides exact ground truth and broad axis coverage, while real-world data captures natural correlations, noise, and stylistic diversity of professional cinematography. As shown in Section~\ref{sec:experiments}, combining both is critical for generalization to in-the-wild captions; the synth-only ablation degrades sharply on paraphrased real-world prompts.
\section{Experiments}
\label{sec:experiments}

We evaluate \name on both a controlled synthetic benchmark and existing camera-conditioned video datasets. Our evaluation focuses on two aspects: (i) adherence to cinematographic intent specified in text prompts, and (ii) robustness of camera–subject alignment during motion. 

\paragraph{Framing metrics.} Standard metrics such as \textit{out-of-frame rate (out-rate)} and \textit{Fréchet distance} on absolute camera do not capture whether a model correctly executes cinematographic instructions. The \textit{out-rate} merely confirms whether a subject is visible, completely ignoring spatial composition; while Fréchet distance evaluates generic distribution similarity without capturing whether a model actually executed a specific stylistic command, such as a precise orientation or a low-angle close-up. We therefore introduce a set of framing-specific metrics that directly measure adherence to intended shot properties. Given a generated trajectory, we compute:
(i)~\textit{F-Ori}: alignment between the camera orientation and desired actor-facing camera direction; 
%
(ii)~\textit{F-RoT}: adherence to the rule-of-thirds composition; 
(iii)~\textit{F-Scale}: conformity with the target shot scale; 
(iv)~\textit{F-Tilt}: accuracy of camera elevation angle; 
(v)~\textit{F-Roll}: accuracy of the Dutch angle.

Each metric is normalized to $[0,1]$, where 1 indicates perfect adherence to the target framing. To evaluate not just static composition but the accurate execution of dynamic camera choreography over time, we compute the average of these five metrics for all the frames of the generated sequence. The mean yields the final \textit{Auteur Score}, an aggregated metric where a score of 1 represents perfect intended cinematographic choreography.

\paragraph{Controlled diagnostic benchmark.} Evaluating camera control models on in-the-wild datasets can mask their true capabilities due to strong dataset biases, such as the prevalence of centered, eye-level shots. To isolate and rigorously evaluate adherence to cinematographic instructions, we construct a \emph{Controlled Diagnostic Benchmark}. The benchmark is designed to evaluate each framing attribute independently. For each metric (e.g., shot scale, orientation, or composition), we generate a set of test sequences where the target attribute is sampled from a uniform distribution over its valid range. This ensures coverage of diverse and challenging configurations, preventing models from exploiting dataset priors. The remaining camera parameters are randomly sampled to maintain variability and avoid trivial solutions. Each test sequence consists of a prompt specifying the desired camera behavior, along with a corresponding ground-truth framing configuration. This allows direct measurement of how well a model follows explicit cinematographic instructions.

\paragraph{Competing methods.}
(i)~\textbf{LAMP}~\citep{LAMP} models camera motion via structured 3D trajectory programs, providing strong geometric control but limited framing awareness. (ii)~\textbf{PulpMotion}~\citep{PulpMotion}, in contrast, introduces a learned screen-space framing signal that improves subject coherence but lacks explicit, controllable cinematographic structure. All methods are evaluated under identical prompts and trajectory settings to ensure fair comparison. We train and test our method on a single NVIDIA A100 GPU. Training takes about 8 hours.
(iii)~Additionally, to contextualize performance on this benchmark, we include a naive static \textbf{Baseline} consisting of a stationary camera with a centered composition observing a non-moving subject. This Baseline approximates the dominant mode of the data distribution and therefore achieves moderate scores on several framing metrics without actively following the prompt. Importantly, this baseline highlights a key challenge in dynamic camera control: while dynamic methods aim to execute complex motion, they may introduce framing instability during movement, leading to subject drift or out-of-frame artifacts. In contrast, the static baseline trivially maintains full subject visibility, resulting in a 0\% out-rate. We include this baseline as a reference point for interpreting the trade-off between motion expressiveness and framing stability.

\paragraph{Main results.}
\Cref{tab:auteurscore} reports the quantitative results on the controlled benchmark. Our method significantly outperforms prior approaches across all framing metrics, demonstrating that modeling camera behavior in a human-centric framing space leads to more accurate execution of cinematographic intent. In particular, the large improvement in F-Ori and F-RoT indicates better alignment between camera orientation and subject composition. Our method also reduces out-of-frame errors (Out-Rate) by a large margin, showing improved stability during motion.

\paragraph{Isolating the representation.}
To isolate the benefit of our actor-relative formulation independently of the model architecture, we compare it against a world-space domain-specific language (DSL) baseline. Because prior world-space DSLs are not directly invertible, we evaluate this by fitting our ground-truth trajectories with a free-form world-space segment DSL. Even when granted significant advantages—specifically, reconstructing from ground truth rather than predicting from language, and receiving perfect first-frame alignment—the world-space representation struggles. Increasing its segment budget from 4 to 8 yields only a 0.280 CamF1 score, compared to 0.623 for our representation. The largest performance gaps occur on the framing axes, demonstrating that world-space segments cannot inherently anchor the camera to a moving actor and are not suitable for sophisticated camera movements.

\begin{table}[!t]
\centering
\small
\caption{\textbf{Framing accuracy on the controlled benchmark.} Mean per-frame framing scores are reported. Rows 5-6 represent an ablation isolating the actor-relative representation by reconstructing ground-truth (GT) trajectories using a world-space DSL.}
\label{tab:auteurscore}

\begin{tabular}{l@{$\;\;$}c@{$\;\;$}c@{$\;\;$}c@{$\;\;$}c@{$\;\;$}c@{$\;\;$}c@{$\;\;$}c@{$\;\;$}c}
\toprule
 & F-Ori & F-RoT & F-Scale & F-Tilt & F-Roll & CamF1 & Auteur-Score & Out-Rate \\
\midrule
Baseline & 0.495 & 0.625 & 0.695 & 0.875 & 0.880 & 0 & - & 0 \\
\midrule
LAMP & 0.451 & 0.257 & 0.729 & 0.832 & 0.745 & 0.112 & 0.603 & 41.0 \\
PulpMotion & \underline{0.514} & 0.351 & 0.712 & \underline{0.891} & \underline{0.839} & 0.091 & \underline{0.661} & 36.2 \\
Ours & \textbf{0.962} & \textbf{0.745} & \textbf{0.888} & \textbf{0.968} & \textbf{0.937} & \textbf{0.623} & \textbf{0.901} & \textbf{5.45} \\
\midrule
World-space, 4-seg (GT recon.) & 0.499 & 0.371 & 0.730 & 0.675 & 0.311 & 0.262 & - & 0 \\
World-space, 8-seg (GT recon.) & 0.500 & \underline{0.372} & \underline{0.733} & 0.674 & 0.313 & \underline{0.280} & - & 0 \\
\bottomrule
\end{tabular}
\end{table}

\paragraph{Generalization on existing benchmarks.}
We further evaluate \name on the PulpMotion benchmark under \emph{pure} and \emph{mixed} settings (see PulpMotion for details).
Results are shown in \Cref{tab:pulpmotion}. On the \emph{pure} split, our method achieves the highest Camera CLaTr score (a contrastive language-trajectory alignment metric introduced by \citep{courant2024et}) of 71.1, while reducing the out-of-frame rate to 1.26\%, compared to 7.00\% for LAMP and over 18\% for PulpMotion variants. This indicates that modeling camera behavior in a human-centric framing space leads to substantially improved subject alignment and stability during motion. While LAMP attains the highest Camera F1 score (a multi-class F1 metric for camera movement translation), this reflects its strength in reproducing geometric trajectories rather than maintaining consistent subject-centric framing, as evidenced by its higher out-of-frame rate.

On the \emph{mixed} split, which introduces more diverse and challenging scenarios, our method maintains an almost zero out-of-frame rate (0.01\%), demonstrating strong robustness across data distributions. In contrast, both PulpMotion variants exhibit significant degradation in out-rate, exceeding 24\%, suggesting instability in maintaining subject visibility under more complex conditions. Although LAMP achieves higher F1 and CLaTr scores, its performance remains limited by its reliance on world-space trajectory modeling, which does not explicitly enforce subject-centric framing constraints.

These results highlight a key distinction between approaches. LAMP models camera motion as a geometric planning problem in world coordinates, which leads to accurate trajectories but weaker control over subject framing. PulpMotion introduces an implicit framing signal through learned latents, improving perceptual quality but lacking explicit controllability. In contrast, \name explicitly models camera behavior as a function of the subject, enabling more reliable framing and significantly improved robustness in maintaining subject visibility. Overall, the results demonstrate that a framing-centric formulation provides a more effective interface for controllable video generation, particularly in scenarios where maintaining consistent subject composition is critical. See supplemental webpage.

\begin{table}[!t]
\centering
\caption{\textbf{PulpMotion benchmark (pure and mixed).} Best/second-best results are in bold/underlined.}
\label{tab:pulpmotion}
\resizebox{0.94\textwidth}{!}{  
\begin{tabular}{l@{$\;\;$}c@{$\;\;$}c@{$\;\;$}c@{$\;\;$}c@{$\quad$}c@{$\;\;$}c@{$\;\;$}c@{$\;\;$}c}
\toprule
 & \multicolumn{4}{c}{\textit{Pure}} & \multicolumn{4}{c}{\textit{Mixed}} \\
\cmidrule(lr){2-5}\cmidrule(lr){6-9}
 & Cam F1 & CLaTr & Out-rate & $FD_{frame}$
 & Cam F1 & CLaTr & Out-rate & $FD_{frame}$ \\
\midrule
LAMP
  & \textbf{99.7} & \underline{69.5} & \underline{7.00} & 35.5
  & \textbf{71.3} & \textbf{59.2}    & \underline{7.02} & 33.57 \\
PulpMotion (DiT)
  & 52.9          & 39.7             & 18.5             & \textbf{4.19}
  & 34.8          & 31.0             & 24.9             & \textbf{5.36} \\
PulpMotion (MAR)
  & 64.8          & 49.3             & 29.1             & \underline{6.80}
  & 38.7          & 39.5             & 39.5             & 8.63 \\
Ours 
  & \underline{83.6} & \textbf{71.1} & \textbf{1.26}    & 8.08
  & \underline{54.0} & \underline{48.5} & \textbf{0.01} & \underline{8.03} \\
\bottomrule
\end{tabular}}
\end{table}

%
%
%
%
%

\paragraph{Ablation studies.}
We evaluate the effect of combining real and synthetic training data. While synthetic data offers perfect ground truth, its captions are linguistically rigid. To simulate in-the-wild queries, we test on a challenging split with LLM-augmented and paraphrased descriptions. Although this augmentation naturally introduces some linguistic ambiguity, it accurately reflects the complexity of real user intent. \Cref{tab:ablation_data} compares the full model with a synthetic-only baseline. The `Translation' metric denotes strict accuracy across all three spatial directions ($t_x, t_y, t_z$). 

Results show that real-world data is critical for grounding diverse narrative intent, preventing severe performance drops in complex parameters like Shot Scale (SS), and making Auteur robust to natural language. Importantly, training with real augmented captions substantially improves performance on real-world paraphrased prompts while introducing only minimal degradation on the synthetic test set, indicating strong generalization across both domains. Both models achieve zero format errors.

\paragraph{Integration across generation paradigms.} To demonstrate the generality of \name, we integrate it with multiple video generation frameworks spanning different conditioning modalities:
\begin{enumerate}[label=(\roman*),leftmargin=*]
\item \textbf{Image-to-Video (I2V).} We interface with VerseCrafter~\citep{versecrafter}, which accepts explicit camera parameters. The predicted 6-DoF camera trajectories from \name are directly used to guide the generation process, enabling controlled viewpoint changes consistent with the actor’s motion.
\item \textbf{Joint human motion and camera generation.} We integrate with Kimodo~\citep{kimodo2026}, which refines coarse motion trajectories into articulated human motion. By combining Kimodo with our predicted camera trajectories, we enable synchronized generation of human motion and camera behavior, allowing rendering from arbitrary, dynamically controlled viewpoints.
\item \textbf{Text-to-Video (T2V).} For models without explicit camera interfaces, such as VACE~\citep{vace}, we convert predicted trajectories into rendered control sequences (e.g., projected actor geometry under the predicted camera). These sequences serve as conditioning signals, enabling implicit control over framing and perspective without architectural modification.
\end{enumerate}

\paragraph{Cross-renderer evaluation.}
To quantitatively evaluate the efficacy of our trajectories when transferred to downstream generators, we measure prompt-adherence using VQAScore (the $p(\text{yes})$ formulation introduced in CHAI ~\citep{lin2026buildingprecisevideolanguage}), which acts as an automatic stand-in for human judgment on whether the rendered video follows the requested camera and framing instructions. Crucially, we compare against a strong baseline—Wan2.2 direct TI2V \citep{wan2025} 5B model using the same prompt but without explicit trajectory guidance—isolating the contribution of \name's trajectory itself.

As shown in \Cref{tab:cross_renderer}, every generator driven by \name's trajectory improves prompt-adherence substantially over the Wan2.2 baseline (achieving $0.47$--$0.50$ vs.\ $0.39$). This directly demonstrates two key properties: (i) independent renderers beyond VerseCrafter—including EPiC \citep{epic} and ReCamMaster\citep{recammaster}—faithfully execute \name's predicted trajectories while preserving output quality, and (ii) the framing improvements are not an artifact of a single renderer, as prompt-adherence consistently rises over the unguided baseline across all architectures. We provide side-by-side video comparisons of these integrations on the supplementary website.

\begin{table*}[!t]
\centering
\begin{minipage}[t]{0.58\linewidth}
\centering
\small
\caption{\textbf{Ablation on Training Data Composition.} We evaluate prediction accuracy on both real and synthetic test sets. Training on real data prevents severe performance drops on complex parameters (e.g., Shot Scale) and robustifies the model to linguistically augmented and paraphrased captions.}
\label{tab:ablation_data}
\vspace*{8pt} 
\resizebox{\linewidth}{!}{
\begin{tabular}{l@{$\;\;$}c@{$\;\;$}c@{$\;\;$}c@{$\;\;$}c}
\toprule
& \multicolumn{2}{c}{\textbf{Real Test Set}} & \multicolumn{2}{c}{\textbf{Synthetic Test Set}} \\
\cmidrule(lr){2-3} \cmidrule(lr){4-5}
\textbf{DSL Component} & \textbf{Synth-Only} & \textbf{Ours} & \textbf{Synth-Only} & \textbf{Ours} \\
\midrule
Translation & 96.4\% & \textbf{97.8\%} & 99.5\% & \textbf{99.8\%} \\
Human Yaw ($\Psi$) & 93.7\% & \textbf{97.5\%} & \textbf{98.8}\% & 98.5\% \\
\midrule
Orientation Angle (OA) & 84.6\% & \textbf{85.1\%} & 98.6\% & 98.6\% \\
Shot Scale (SS) & 60.1\% & \textbf{85.8\%} & 95.0\% & \textbf{95.2\%} \\
Camera Level (CL) & 92.8\% & \textbf{93.2\%} & \textbf{97.1\%} & 96.7\% \\
Look-at Level (LL) & 90.2\% & \textbf{90.5\%} & \textbf{96.8\%} & 96.3\% \\
Framing Offset (FO) & 86.1\% & \textbf{88.6\%} & \textbf{99.1\%} & 97.9\% \\
Dutch Angle (DA) & 98.9\% & \textbf{99.3\%} & \textbf{100.0\%} & 99.8\% \\
\bottomrule
\end{tabular}
}
\end{minipage}\hfill
\begin{minipage}[t]{0.38\linewidth}
\centering
\small
\caption{\textbf{Cross-Renderer Prompt Adherence.} VQAScore ($p(\text{yes})$) evaluating how faithfully the rendered video executes the requested camera/framing instruction. }
\label{tab:cross_renderer}
\vspace*{8pt}
\resizebox{\linewidth}{!}{
\begin{tabular}{lc}
\toprule
\textbf{Configuration} & \textbf{VQAScore} ($p(\text{yes})$) \\
\midrule
Baseline (Wan2.2) & 0.39 \\
\midrule
Ours + VerseCrafter & \textbf{0.50} \\
Ours + ReCamMaster & \underline{0.49} \\
Ours + EPiC & 0.47 \\
\bottomrule
\end{tabular}
}
\end{minipage}
\vspace{-10pt}
\end{table*}
%



\paragraph{User study.} 
We conducted a blind A/B test to validate perceptual effectiveness. Across 10 scenarios, paired videos used identical prompts, images, and rendering settings—differing solely in the predicted camera trajectory. From 225 pairwise judgments (24 participants), users chose which video looked more professional (quality) and better followed instructions (adherence). \name achieved overwhelming win rates against prior works: \textbf{82.7\%} (quality) and \textbf{80.4\%} (adherence) against LAMP \citep{LAMP}; and \textbf{81.3\%} (quality) and \textbf{83.1\%} (adherence) against PulpMotion \citep{PulpMotion}. This confirms our trajectory-level gains yield perceptually superior videos. We provide an additional automatic perceptual evaluation in the (\Cref{subsec:supp_vbench}) which further supports findings.
\section{Conclusion}
We presented \name, a language-driven framework for cinematographic camera framing in human-centric video generation. \name grounds camera control in actor-relative framing rather than world-space trajectories, and expresses it through a deterministically decodable DSL generated by a fine-tuned LLM. By formulating camera behavior as a conditional response to human motion, our approach enables more consistent subject alignment, interpretable control, and improved robustness across diverse scenarios. Extensive experiments demonstrate that this framing-centric formulation leads to significant gains in both controllability and stability compared to prior trajectory-based methods. Future work includes extending the framework to multi-actor interactions, richer interaction dynamics, and integrating learned policies more tightly with video generation architectures.


\textbf{Limitations.} Our DSL keyframe representation assumes that cinematographic intent can be expressed as a sparse sequence of discrete states; highly continuous or reactive shot styles (e.g., handheld verité, continuous reframing during action) may be underrepresented in the vocabulary. We only model actor trajectories without fine articulation, augmenting \name with advanced human motion synthesis capabilities is an interesting future direction. Finally, while \name reliably controls framing, the perceptual quality of the final video remains largely determined by the downstream generator.

\textbf{Broader Impact.} \name lowers the barrier to professional-quality cinematography for creators without formal filmmaking training, with potential applications in accessibility, independent production, and education. At the same time, the same capability could be misused to produce more convincing synthetic media; we encourage downstream deployments to pair \name with provenance and watermarking tools.
\newpage
{
\small
\bibliographystyle{ieeenat_fullname}
\bibliography{main}
}
\newpage
\renewcommand\thesection{\Alph{section}}
\renewcommand{\thefigure}{A.\arabic{figure}}
\renewcommand{\thetable}{A.\arabic{table}}
\setcounter{section}{0}
\setcounter{figure}{0}
\setcounter{table}{0}

\section{Appendix}
\subsection{Cinematography DSL}
\label{app:dsl}
 We define a discrete DSL as a quantized, human-readable version of the human centric camera parameter space. As shown in Table~\ref{tab:axes}, each axis is equipped with a finite vocabulary of cinematographically motivated tokens.
 
\begin{table}[h]
\centering
\caption{\textbf{Camera axes: continuous domains, discrete DSL vocabularies, and token-to-scalar mappings.} Each axis $a \in \mathcal{A}$ has a continuous domain $\mathcal{K}_a$, a finite vocabulary $\mathcal{V}_a$, and a deterministic embedding $\iota_a : \mathcal{V}_a \hookrightarrow \mathcal{K}_a$. Scalar values are expressed in meters (CL, LL), unitless fractions (SS, FO), or degrees (OA, DA). Shot-scale scalars denote the fraction of frame height occupied by the actor's body height $h_t$.}
\label{tab:axes}
\small
\begin{tabular}{@{}llllp{5.8cm}@{}}
\toprule
Axis & Symbol & $\mathcal{K}_a$ & Geometry & $\mathcal{V}_a$ (token $\to$ scalar via $\iota_a$) \\
\midrule
Camera Level & CL & $\mathbb{R}_{\geq 0}$ & $\mathbb{R}$ &
  \textsc{Ground}~$\to 0.1$\,m, \;
  \textsc{Low}~$\to 0.6$\,m, \;
  \textsc{Eye}~$\to 1.6$\,m, \;
  \textsc{High}~$\to 2.5$\,m, \;
  \textsc{Overhead}~$\to 4.0$\,m \\[4pt]
Look-at Level & LL & $[0,\, h_t]$ & $\mathbb{R}$ &
  \textsc{Feet}~$\to 0.0$, \;
  \textsc{Knees}~$\to 0.25\,h_t$, \;
  \textsc{Waist}~$\to 0.50\,h_t$, \;
  \textsc{Chest}~$\to 0.75\,h_t$, \;
  \textsc{Eyes}~$\to 0.95\,h_t$ \\[4pt]
Shot Scale & SS & $(0,\, 1]$ & $\mathbb{R}$ &
  \textsc{ECU}~$\to 0.10$, \;
  \textsc{CU}~$\to 0.20$, \;
  \textsc{MCU}~$\to 0.35$, \;
  \textsc{MS}~$\to 0.50$, \;
  \textsc{MLS}~$\to 0.65$, \;
  \textsc{LS}~$\to 0.85$, \;
  \textsc{EWS}~$\to 1.00$ \\[4pt]
Framing Offset & FO & $[-\tfrac{1}{2},\, +\tfrac{1}{2}]$ & $\mathbb{R}$ &
  \textsc{FarLeft}~$\to -0.35$, \;
  \textsc{Left}~$\to -0.17$, \;
  \textsc{Center}~$\to 0.0$, \;
  \textsc{Right}~$\to +0.17$, \;
  \textsc{FarRight}~$\to +0.35$ \\[4pt]
Orientation Angle & OA & $\mathbb{S}^1$ & $\mathbb{S}^1$ &
  \textsc{Front}~$\to 0°$, \;
  \textsc{FrontL}~$\to -45°$, \;
  \textsc{SideL}~$\to -90°$, \;
  \textsc{BackL}~$\to -135°$, \;
  \textsc{Back}~$\to \pm180°$, \;
  \textsc{BackR}~$\to +135°$, \;
  \textsc{SideR}~$\to +90°$, \;
  \textsc{FrontR}~$\to +45°$ \\[4pt]
Dutch Angle & DA & $\mathbb{S}^1$ & $\mathbb{S}^1$ &
  \textsc{None}~$\to 0°$, \;
  \textsc{SlightL}~$\to -10°$, \;
  \textsc{SlightR}~$\to +10°$, \;
  \textsc{StrongL}~$\to -25°$, \;
  \textsc{StrongR}~$\to +25°$ \\
\bottomrule
\end{tabular}
\end{table}

By changing the camera axes properties at different keyframes, our DSL captures common camera behaviour. For example, the orientation determines whether a particular tracking motion is \emph{tail, lead,} or \emph{side}. A change in shot scale results in a \emph{dolly in} or \emph{out}. The camera and look-at levels together determine tilt behaviour. A change in roll results in the classic \emph{duth angle} effect. 

\subsection{From DSL to 6-DoF Camera Trajectories}
\label{app:interp}

We now describe how a sparse keyframe sequence in $\mathcal{K}$ becomes a dense $\mathrm{SE}(3)$ trajectory.

\paragraph{Geometric decoder $\Phi$.}
Given a dense state $\mathbf{k}_t \in \mathcal{K}$, an actor state $a_t$, and fixed intrinsics Field-of-View and Aspect Ratio $(\mathrm{FOV}, \mathrm{AR})$, the decoder $\Phi(\mathbf{k}_t; a_t, \mathrm{FOV}, \mathrm{AR}) \in \mathrm{SE}(3)$ is defined as follows. The image-plane span occupied by the actor's body of height $h_t$ at shot scale $\mathrm{SS}$ is
\begin{equation}
H \;=\; {h_t}/{\mathrm{SS}}, \qquad W \;=\; \mathrm{AR} \cdot H,
\end{equation}
giving an actor-camera distance 
\begin{equation}
d \;=\; \frac{H}{2 \tan(\mathrm{FOV}/2)},
\end{equation}
and a lateral framing shift $S_f = \mathrm{FO} \cdot W$ in the world. The global azimuth in the ground plane is $\theta = \mathrm{OA} + \psi_t$. 
The camera position $\mathbf{T}_t = (T_t^x, T_t^y, T_t^z) \in \mathbb{R}^3$ is
\begin{align}
T_t^x &= \mathbf{p}_t^x \;+\; d\cos\theta \;-\; S_f\sin\theta, \\
T_t^y &= \mathbf{p}_t^y \;+\; d\sin\theta \;+\; S_f\cos\theta, \\
T_t^z &= \mathrm{CL}.
\end{align}
The orientation $\mathbf{R}_t \in \mathrm{SO}(3)$ is determined by the convention that the optical axis points from $\mathbf{T}_t$ toward the look-at target $\mathbf{q}_t = (\mathbf{p}_t^x, \mathbf{p}_t^y, \mathrm{LL})$, modulo a Dutch-angle roll:
\begin{equation}
\mathbf{R}_t \;=\; \mathrm{LookAt}\!\big(\mathbf{T}_t \to \mathbf{q}_t,\, \hat{\mathbf{z}}\big) \cdot \mathbf{R}_z(\mathrm{DA}),
\end{equation}
where $\mathrm{LookAt}$ produces the rotation aligning the camera's $-\hat{\mathbf{z}}_{\text{cam}}$ with $\mathbf{q}_t - \mathbf{T}_t$ and using $\hat{\mathbf{z}}$ as up reference. Equivalently, in Euler-angle form (yaw–pitch–roll, intrinsic),
\begin{equation}
\mathrm{Yaw} = \theta + \pi, \qquad \mathrm{Pitch} = \mathrm{atan2}(\mathrm{LL} - \mathrm{CL},\; d), \qquad \mathrm{Roll} = \mathrm{DA}.
\end{equation}

\paragraph{Domain restrictions.}
$\Phi$ is well-defined on the open subset $\mathcal{K}^\circ \subset \mathcal{K}$ where $\mathrm{SS} \in (\epsilon, 1]$ for some $\epsilon > 0$ (bounding $d$ from above) and where $(\mathrm{LL}, \mathrm{CL}, d)$ are not simultaneously zero (avoiding pitch degeneracy). Both DSL vocabularies and dataset construction enforce $\mathcal{K}^\circ$.

\paragraph{Axis-aware interpolation in $\mathcal{K}$.}
Note that we deliberately interpolate in $\mathcal{K}$ rather than $\mathrm{SE}(3)$: cinematographically meaningful axes vary smoothly under their natural metrics (e.g., a linear ramp in SS produces a perceptually uniform zoom), whereas $\mathrm{SE}(3)$ interpolation entangles rotation and translation in ways that violate framing intent. 
\subsection{Ablation on Actor-First Factorization}
\label{sec:actor}
As formulated in the main text, \name predicts the actor's motion before the camera's response. This actor-first factorization is a deliberate design choice driven by causality and utility rather than pure predictive accuracy. 

As shown in \Cref{tab:factorization}, alternative autoregressive orderings—such as generating the camera first or generating both jointly—perform comparably across all axes on the synthetic test set, with no single formulation emerging as a strict winner. 

We adopt the actor-first formulation for two fundamental reasons. First, the ordering is strictly dictated by the actor-relative pipeline: because the camera is parameterized relative to the actor, generating camera tokens first is inherently ill-posed. Without an established actor trajectory, there is nothing to anchor the camera to, meaning a standalone camera program cannot be decoded into a valid world-space trajectory. Second, this factorization enables practical downstream control. It allows the camera generation module to be conditioned on an externally provided or user-defined actor sequence, cleanly separating subject motion from camera response in a way that matches real-world cinematography. The empirical results in \Cref{tab:factorization} confirm that we incur no performance penalty by adopting this necessary, causally grounded factorization.

\begin{table}[h]
\centering
\small
\caption{\textbf{Effect of Factorization Ordering.} Prediction accuracy (\%) on the synthetic test set across different autoregressive generation orders. The required actor-first factorization performs on par with alternative orderings.}
\label{tab:factorization}
\begin{tabular}{l@{$\;\;$}c@{$\;\;$}c@{$\;\;$}c}
\toprule
\textbf{DSL Component} & \textbf{Camera-First} & \textbf{Joint} & \textbf{Ours (Actor-first)} \\
\midrule
Translation & 99.6 & 99.6 & \textbf{99.8} \\
Human Yaw ($\Psi$) & \textbf{99.3} & \textbf{99.3} & 98.5 \\
\midrule
Orientation Angle (OA) & \textbf{98.7} & 98.6 & 98.6 \\
Shot Scale (SS) & 94.1 & 93.0 & \textbf{95.2} \\
Camera Level (CL) & \textbf{98.7} & 98.2 & 96.7 \\
Look-at Level (LL) & \textbf{98.5} & 98.2 & 96.3 \\
Framing Offset (FO) & 93.3 & 93.0 & \textbf{97.9} \\
Dutch Angle (DA) & 96.8 & 96.7 & \textbf{99.8} \\
\bottomrule
\end{tabular}
\end{table}

\subsection{Automatic Perceptual Evaluation}
\label{subsec:supp_vbench}

To complement the human user study with an automatic video-level measure, we evaluated the generated videos using VBench \citep{huang2023vbenchcomprehensivebenchmarksuite}. The evaluation was conducted on the same set of scenarios used in our human trials, ensuring the downstream generator and rendering settings remained strictly identical across all methods. 

As reported in \Cref{tab:vbench}, \name leads across most evaluated dimensions. Notably, our method shows strong improvements in Subject Consistency, Dynamic Degree, and overall Aesthetic Quality. These quantitative video-level results align closely with the user study, confirming that the framing and trajectory improvements of \name directly yield higher-quality final video outputs.

\begin{table}[h]
\centering
\small
\caption{\textbf{Automatic Perceptual Evaluation (VBench).} Video-level quality metrics (\%) for generated videos. \name achieves the highest scores across the majority of perceptual dimensions.}
\label{tab:vbench}
\begin{tabular}{l@{$\;\;$}c@{$\;\;$}c@{$\;\;$}c}
\toprule
\textbf{Metric} & \textbf{PulpMotion} & \textbf{LAMP} & \textbf{Ours} \\
\midrule
Subject Consistency & 87.72 & 89.50 & \textbf{91.23} \\
Dynamic Degree & 81.82 & 54.55 & \textbf{90.91} \\
Background Consistency & 90.86 & 91.37 & \textbf{93.01} \\
Motion Smoothness & 98.32 & \textbf{99.08} & 98.99 \\
Aesthetic Quality & 59.88 & 59.35 & \textbf{63.11} \\
Image Quality & 69.36 & \textbf{69.65} & 68.71 \\
\bottomrule
\end{tabular}
\end{table}

\subsection{Training details}
\label{app:training}
We fine-tune the \texttt{Qwen2.5-VL-7B-Instruct} model using parameter-efficient LoRA adaptation with rank 128 and $\alpha=128$. Training is performed for a single epoch over approximately 300K image-text samples. We use a global batch size of 512 with gradient accumulation, BF16 mixed-precision training, cosine learning rate scheduling, and a peak learning rate of $1\times10^{-4}$. Following prior efficient multimodal fine-tuning practices, the vision encoder and LLM backbone are frozen while only the multimodal merger layers and LoRA adapters are optimized.

\subsection{Dataset}
\label{app:dataset}
This appendix expands on the brief description in \Cref{sec:datasets}. We provide the full procedural-synthesis pipeline, the four-stage real-world annotation pipeline, dataset statistics, and a comparison to existing camera-trajectory datasets.

\subsubsection{Procedural Synthesis}
\label{app:dataset-procedural}
The procedural split consists of $N_{\mathrm{proc}}$ tuples $(\ell, \mathcal{P}_{\mathrm{act}}, \mathcal{P}_{\mathrm{cam}})$ synthesized in three stages.

\paragraph{Motion sampling.}
We sample human motion clips from the SOMA library~\citep{saito2026soma}, covering common locomotion (walking, running, turning), gestural actions (pointing, reaching, sitting), and stationary poses. Each clip is a sequence $\{a_t\}_{t=1}^T$ of pelvis position $\mathbf{p}_t$, yaw $\psi_t$, and a clip-level body height $h_t$ drawn from a realistic human-height distribution.

\paragraph{Camera program sampling.}
For each motion clip we independently sample a camera DSL program $\mathcal{P}_{\mathrm{cam}}$ from our grammar with axis-stratified priors. Concretely, we (i) sample the number of keyframes $K \sim \mathrm{Uniform}\{1, \ldots, K_{\max}\}$, (ii) sample keyframe indices $\tau_1 < \cdots < \tau_K$ uniformly within $[1, T]$, and (iii) for each $\tau_k$ sample a partial assignment $\delta_k$ by 
first drawing the cardinality $|\mathrm{dom}(\delta_k)|$ from a distribution that explicitly upweights multi-axis transitions, then drawing token values uniformly from each axis vocabulary. This sampling strategy ensures the LLM sees sufficient examples of rare combinations such as simultaneous five- or six-axis changes, which are vanishingly rare in real-world footage (\Cref{tab:dataset_stats}, left).

\paragraph{Decoding and caption generation.}
Each $(\mathcal{P}_{\mathrm{act}}, \mathcal{P}_{\mathrm{cam}})$ pair is decoded via the deterministic pipeline of \Cref{sec:representation} and Appendix~\ref{app:interp}: carry-forward decoding, embedding via $\varphi$, axis-aware spline interpolation, and the geometric decoder $\Phi$ produce a ground-truth dense actor state $\mathbf{a}_{1:T}$ and 6-DoF camera trajectory $\{(\mathbf{R}_t, \mathbf{t}_t)\}_{t=1}^T$. Captions are generated by structured templates that verbalize the sampled motion and framing parameters: e.g., a clip with $\mathrm{SS}{=}\textsc{MS}$, $\mathrm{OA}{=}\textsc{Front}$, $\mathrm{CL}{=}\textsc{Eye}$ over a forward-walking actor yields the caption \emph{``A person walks forward; medium shot, frontal, eye-level.''} While these captions are linguistically rigid, they provide unambiguous supervision for the LLM to associate cinematic vocabulary with DSL outputs; we ablate the importance of augmenting this with real-world captions in \Cref{sec:experiments}.

\subsubsection{Real-World Pipeline}
\label{app:dataset-realworld}
The real-world split contains $N_{\mathrm{real}}$ tuples mined from CondensedMovies~\citep{condensedMovies}. Each clip is processed through a four-stage pipeline.

\paragraph{(i) 3D reconstruction.}
Following TRAM~\citep{tram}, a joint human-and-camera estimator recovers metric-scale global camera extrinsics $\{(\mathbf{R}_t, \mathbf{t}_t)\}_{t=1}^T$ and per-frame SOMA body parameters from monocular video. Clips for which TRAM fails to converge or returns low-confidence estimates are discarded.

\paragraph{(ii) Projection to $\mathcal{K}_{\mathrm{cam}}$.} The recovered camera and actor trajectories are projected into the human-centric parameter space defined in Section~\ref{sec:representation}: for each frame $t$, we compute the six camera axis values $(\mathrm{CL}, \mathrm{LL}, \mathrm{SS}, \mathrm{FO}, \mathrm{OA}, \mathrm{DA})_t$ from the world-space pose relative to the body frame $\mathcal{F}_t$. This yields a dense continuous trajectory in $\mathcal{K}_{\mathrm{cam}}$ for each clip.

\paragraph{(iii) Motion tagging.}
We extract a sparse keyframe program from the dense trajectory using a per-axis change-point detector. For each axis $a$, we identify frames at which the axis value undergoes a transition larger than a threshold $\Delta_a$ (calibrated per axis to reflect cinematographic significance, e.g., a change of one shot-scale category for SS, or a $30^\circ$ azimuth change for OA). At each detected change point, the new axis value is snapped to the nearest token in $\mathcal{V}_a$ via the inverse of the per-axis lookup $\varphi_a$. The resulting keyframes form $\mathcal{P}_{\mathrm{cam}}$. The same procedure applied to the actor trajectory yields $\mathcal{P}_{\mathrm{act}}$.

\paragraph{(iv) Captioning.} We obtain a free-form natural-language caption $\ell$ for each clip using AuroraCap~\citep{chai2024auroracap}. To improve linguistic diversity and robustness to in-the-wild user prompts, we additionally generate paraphrased variants by prompt rewriting with Gemini~\citep{geminiteam2024gemini15}. Each clip is associated with multiple caption variants during training, sampled uniformly per epoch.

Each successfully processed clip contributes one tuple $(\ell, \mathcal{P}_{\mathrm{act}}, \mathcal{P}_{\mathrm{cam}})$ to the real-world split.



\subsubsection{Statistics}
\label{app:dataset-stats}

\paragraph{Real-world split.}
\Cref{tab:dataset_stats} reports the distribution of camera-and actor-field changes per sequence in the single human subset of real-world split. Most sequences are static or involve one-to-two camera-field changes per clip; multi-axis transitions (three or more axes changing simultaneously) form a long but well-represented tail. Among individual axes, orientation and shot-scale changes are the most frequent camera events, consistent with standard cinematographic practice that uses repositioning and re-scaling as the primary shot-transition mechanisms~\citep{mascelli1965five, Bordwell2020}. Dutch-angle changes are rare in mainstream cinema, which is reflected in the low count for that axis; the procedural split compensates by upweighting Dutch-angle samples during synthesis.

\begin{table}[!h]
\centering
\caption{\textbf{Real-world split statistics.} \emph{Left:} number of camera-field changes per sequence; most sequences are static or involve one--two field changes. \emph{Center:} per-field change frequency; orientation and shot-scale (depth) changes dominate, consistent with standard cinematographic practice~\citep{mascelli1965five, Bordwell2020}. \emph{Right:} human-field changes per segment.
}
\label{tab:dataset_stats}
\smallskip
\small
\begin{minipage}[t]{0.30\linewidth}
    \centering
    \begin{tabular}{lr}
        \toprule
        \textbf{\# Changes} & \textbf{Count} \\
        \midrule
        Static (0)  & 14{,}639 \\
        1-field     &  1{,}910 \\
        2-field     &  1{,}534 \\
        3-field     &    917 \\
        4-field     &    440 \\
        5-field     &    158 \\
        6-field     &      8 \\
        \bottomrule
    \end{tabular}
\end{minipage}
\hfill
\begin{minipage}[t]{0.33\linewidth}
    \centering
    \begin{tabular}{lr}
        \toprule
        \textbf{Camera Field} & \textbf{Changed} \\
        \midrule
        Orientation     & 2{,}527 \\
        Shot scale      & 2{,}403 \\
        Framing offset  & 2{,}281 \\
        Look-at level   & 1{,}820 \\
        Camera level    & 1{,}184 \\
        Dutch angle     &    112 \\
        \bottomrule
    \end{tabular}
\end{minipage}
\hfill
\begin{minipage}[t]{0.28\linewidth}
    \centering
    \begin{tabular}{lr}
        \toprule
        \textbf{\# Changes} & \textbf{Count} \\
        \midrule
        Static (0)  & 25{,}682 \\
        1-field     &  5{,}438 \\
        2-field     &  4{,}085 \\
        3-field     &  3{,}171 \\
        4-field     &    836 \\
        \bottomrule
    \end{tabular}
\end{minipage}
\end{table}

\paragraph{Comparison to existing datasets.}
\Cref{tab:dataset_comparison} compares the \name dataset to prior camera-trajectory datasets along several axes: support for camera and human annotation, multi-human scenes, frame and sample counts, and average caption length. While prior datasets such as DataDoP~\citep{zhang2025gendop}, E.T.~\citep{courant2024et}, and PulpMotion~\citep{PulpMotion} are larger in raw scale, \name is the first to combine (i) explicit human annotation, (ii) multi-human support, and (iii) DSL-aligned camera framing labels in a single corpus. Our contribution emphasizes annotation quality and structural alignment with cinematographic conventions rather than raw scale; this design choice is validated by the strong cross-dataset generalization reported in \Cref{sec:experiments}.

\begin{table}[!h]
\centering
\caption{\textbf{Dataset comparison.} Comparison of various datasets with a focus on camera and human attributes. Our proposed dataset is shown at the bottom.}
\footnotesize
\label{tab:dataset_comparison}
\begin{tabular}{l@{$\;\;$}c@{$\;\;$}c@{$\;\;$}c@{$\;\;$}c@{$\;\;$}c@{$\quad$}c@{$\;\;$}c@{$\quad$}c@{$\;\;$}c@{$\quad$}c@{$\;\;$}c}
\toprule
 & & & & & & \multicolumn{2}{c}{\textit{Vocabulary}} & \multicolumn{2}{c}{\textit{Avg. Cap. Len.}} & & \\
\cmidrule(lr){7-8}\cmidrule(lr){9-10}
Dataset & Camera & Human & Multi-Human &\#Frames & \#Samples & Cam. & Hum. & Cam. & Hum.  \\
\midrule
DataDoP \citep{zhang2025gendop} & \checkmark & $\times$ & $\times$ & 11M & 29K &8698 & $\times$ &86.2 & $\times$ \\
E.T. \citep{courant2024et}      & \checkmark & \checkmark & $\times$ & 11M & 115K & 906 & $\times$ & 11.58 & $\times$  \\
PulpMotion \citep{PulpMotion}   & \checkmark & \checkmark& $\times$  & 22M & 193K & 1002 & 4007& 11.72 & 13.92 \\
\midrule
Ours                           & \checkmark & \checkmark&\checkmark&2.7M & 34K & 3264 & 5667&205.8 &120.1  \\
\bottomrule
\end{tabular}
\end{table}

\paragraph{Combined corpus.}
The full training set comprises $N_{\mathrm{proc}}$ procedural and $N_{\mathrm{real}}$ real-world tuples for a total of $34{,}000$ aligned samples. The two splits are deliberately complementary: procedural data ensures broad axis coverage and exact ground truth for supervision of rare combinations, while real-world data exposes the model to the correlations, noise, and stylistic diversity of professional cinematography. The ablation in Section~\ref{sec:experiments} (Table~3) confirms that combining the two is critical. A synth-only model degrades sharply on paraphrased real-world prompts, particularly on linguistically-mediated axes such as shot scale where natural 
captions are most ambiguous.

Overall, we view \name as a step toward more transparent and controllable generative systems, where structured representations can facilitate both creative empowerment and responsible oversight.

\newpage
\end{document}